\documentclass{article} 
\usepackage{nips07submit_e,times}
\usepackage{amsmath,amssymb,amsthm}
\usepackage{algorithm,algorithmic}
\usepackage{graphicx}
\usepackage{multirow}

\newtheorem{theorem}{Theorem}

\def\R{\mathbb{R}}

\def\X{\mathcal{X}}

\def\explanatory{explanatory}
\def\calP{\mathcal{P}}
\newcommand{\changeto}[2]{#2}
\newcommand{\mbf}[1]{\mathbf{#1}}

\newcommand{\Rg}[2]{R_{#1,#2}}
\newcommand{\nhedge}{NormalHedge}
\newcommand{\ls}[2]{\ell_{#1,#2}}
\newcommand{\pr}[2]{w_{#1,#2}}

\newcommand{\resample}{\mathsf{Resample}}

\newcommand{\Ev}{\mathcal{A}}

\newcommand{\calN}{\mathcal{N}}

\def\Explanatory{Explanatory}

\author{{\bf Kamalika Chaudhuri} \\ 
ITA, UC San Diego\\
{\tt kamalika@soe.ucsd.edu}\\
\And Yoav Freund \\
CSE Dept., UC San Diego \\
{\tt yfreund@ucsd.edu}\\
\And Daniel Hsu\\
CSE Dept., UC San Diego\\
{\tt djhsu@cs.ucsd.edu}}

\title{Tracking Using an \Explanatory\ Framework}

\begin{document} 

\maketitle 

\begin{abstract} 

  We study the tracking problem, namely, estimating the hidden state of an
  object over time, from unreliable and noisy measurements.
  The standard framework for the tracking problem is the generative
  framework, which is the basis of solutions such as the Bayesian algorithm
  and its approximation, the particle filters.
  However, the problem with these solutions is that they are very sensitive
  to model mismatches.

  In this paper, motivated by online learning, we introduce a new framework
  -- an {\em explanatory} framework -- for tracking.
  We provide an efficient tracking algorithm for this framework.
  We provide experimental results comparing our algorithm to the Bayesian
  algorithm on simulated data.
  Our experiments show that when there are slight model mismatches, our
  algorithm vastly outperforms the Bayesian algorithm.

\end{abstract} 
 
\section{Introduction}

We study the tracking problem, which has numerous applications in AI, control and finance. 
In tracking, we are given noisy measurements over time, and the problem is to
 estimate the hidden state of an object. The challenge is to do this reliably, by combining measurements from multiple time steps and prior knowledge about the state dynamics, and the goal of tracking is to produce estimates that are as close to the true states as possible.

The most popular solutions to the tracking problem are the Kalman
filter~\cite{Kal60}, the particle filter~\cite{DdFG01}, and their numerous
extensions and variations (\emph{e.g.}~\cite{IB98,vdMDdFW00}), which are 
based on a generative framework for the tracking
problem. Suppose we want to track the state $x_t$ of an object at time $t$,
given only measurement vectors $M(\cdot,t')$ \changeto{at time}{for times} $t' \leq t$. In the generative
approach, we think of the state $X(t)$ and measurements $M(\cdot,t)$ as random
variables. We represent our knowledge regarding the dynamics of the
states using the transition process $\Pr(X(t)|X(t-1))$ and our knowledge
regarding the (noisy) relationship between the states and the observations
by the measurement process $\Pr(M(\cdot,t)|X(t))$. 
Then, given only the observations, the goal of tracking is to estimate the
hidden state sequence $(x_1,x_2,\ldots)$. This is done by calculating the likelihood of each
state sequence and then using as the estimate either the sequence with
the highest posterior probability (maximum a posteriori, or MAP) or the
expected value of the state with respect to the posterior distribution (the
Bayesian algorithm). In practice, one uses particle filters, which are an approximation to the Bayesian algorithm.

The problem with the generative framework is that in practice, it is very
difficult to precisely determine the distributions of the measurements.
Moreover, the Bayesian algorithm is very sensitive to
model mismatches, \changeto{and}{so}
using a model which is slightly different from the model generating the
measurements can lead to a large divergence between the estimated states and the true states.

To address this, we introduce an online-learning-based framework for tracking.
In our framework, called the {\em \explanatory} framework, we are given a
set of state sequences or paths in the state space; but instead of assuming
that the observations are \changeto{generated}{{\em generated}} by a
measurement model from \changeto{{\em a path in this set}}{a path in this
set}, we think of each path as a mechanism for {\em explaining} the
observations.
We emphasize that this is done regardless of how the observations are generated.
Suppose \changeto{that the}{a} path
\changeto{$x^1,x^2,\ldots$}{$(x_1,x_2,\ldots)$} is proposed as an
explanation of the observations
\changeto{$M(\cdot,1),M(\cdot,2),\ldots$}{$(M(\cdot,1),M(\cdot,2),\ldots)$}.
\changeto{Then, we}{We} measure the quality of this \explanatory\ path
using a predefined {\em loss function}, which depends only on the
measurements (and not on the hidden true state).
The tracking algorithm selects its own \explanatory\ path by taking a
weighted average of the best \explanatory\ paths according the past
observations.
The theoretical guarantee we provide is that the loss of the \explanatory\
path generated in this online way by the tracking algorithm is close to
that of the \explanatory\ path with the minimum \changeto{loss, where}{such
loss; here,} the loss is measured according to the loss function supplied
to the algorithm.
Such guarantees are analogous to competitive analysis used in online learning~\cite{CL06, FS97, LW94}, and it is important to note that such guarantees hold uniformly for {\em any} sequence of observations, regardless of any probabilistic assumptions. 

Our next contribution is to provide an online-learning-based algorithm for tracking in the
\explanatory\ framework. Our algorithm is based on \nhedge~\cite{Anon09}, which is a general online learning algorithm. \nhedge\ can be instantiated with {\em any
loss function}. When supplied with a bounded loss function, it is guaranteed to produce a path with loss close to
that of the path with the minimum loss, from a set of candidate paths. As it is inefficient to directly apply \nhedge\ to tracking,
we derive a Sequential \changeto{Monte-Carlo-based}{Monte Carlo}
approximation to \nhedge, and we show that this approximation is efficient.

To demonstrate the robustness of our tracking algorithm,
we perform simulations on a simple one-dimensional tracking problem. 
We evaluate tracking performance by measuring the average distance between
the states estimated by the algorithms, and the true hidden states. We instantiate our algorithm with a simple clipping loss function. Our
simulations show that our algorithm consistently outperforms the Bayesian algorithm, under high measurement noise, and a wide range of levels of model mismatch.

We note that Bayesian algorithm can also be interpreted in the
\explanatory\ framework. In particular, if the loss of a path
is the negative log-likelihood (the log-loss) under some measurement model,
then, the Bayesian algorithm can be shown to
produce a path with log-loss close to
that of the path with the minimum log-loss.
One may be tempted to think that our tracking solution follows the same
approach; however, the point of our paper is that one can use
loss functions that are different from log-loss, and in particular, we show
a scenario \changeto{where}{in which} using other loss functions produces
better tracking performance than the Bayesian algorithm (or its
approximations).

The rest of the paper is organized as follows.
In Section 2, we explain in detail our explanatory model for tracking.
In Section 3, we present \changeto{Normalhedge.}{NormalHedge,
on which our tracking algorithm is based}.
In Section 4, we provide our \changeto{actual tracking algorithm}{tracking
algorithm}.
\changeto{Section 5 presents some experiments that compare our algorithm
with the Bayesian algorithm on simulated data.}{Section 5 presents the
experimental comparison of our algorithm with the Bayesian algorithm.}
\changeto{Finally, in Section 6, we report on our experiments with
face-tracking.}{Finally, we discuss related work in Section 6.}

The detailed bounds and proofs for NormalHedge are provided in the
supplementary material.
We feel that the algorithm NormalHedge may be of more general interest, and
hence these details for NormalHedge have been submitted to NIPS in a
companion paper.  

\section{\changeto{Our Framework: \Explanatory\ Framework}{The \explanatory\
framework for tracking}}
\label{sec:explanation}

In this section, we describe in more detail the setup of the tracking problem, and the
\explanatory\ framework for tracking. In tracking, at each
time $t$, we are given as input, measurements (or observations) $M(\cdot, t)$, and the goal
is to estimate the hidden state of an object using these measurements, and our prior knowledge about the state dynamics.

In the \explanatory\ framework, we are given a set $\calP$ of paths
(\changeto{or }{}sequences) over the state space $\X \subset \R^n$.
At each time $t$, we assign to each path in $\calP$ a loss function $\ell$.
The loss function has two parts\changeto{ --}{:} a dynamics loss
$\ell_d$\changeto{,}{} and an observation loss $\ell_o$.

The dynamics loss $\ell_d$ \changeto{encapsulates}{captures} our
\changeto{prior}{} knowledge about the state dynamics.
For simplicity, we use a dynamics loss $\ell_d$\changeto{,
which}{ that} can be written as\changeto{:}{}
\[ \ell_d(\mbf{p}) = \sum_t \ell_d(x_t, x_{t-1}) \]
for a path $\mbf{p} = (x_1, x_2, \ldots)$.
In other words, the dynamics loss at time $t$ depends only on the states at
time $t$ and $t-1$.
A common way to express our knowledge about the dynamics is in terms of a
dynamics function $F$, \changeto{such that}{defined so that} \changeto{we are
interested in paths in which $x_{t} \changeto{\approxeq}{\approx}
F(x_{t-1})$.}{paths with $x_t \approx F(x_{t-1})$ will have small dynamics
loss.}

For example, consider an object moving
with a constant velocity\changeto{; here, if}{. Here, if} the state $x_t = ( p, v)$, where $p$ is
the position and $v$ is the velocity, then we would be interested in paths
in which $x_{t} \changeto{\approxeq}{\approx} x_{t-1} + (v, 0)$. In these
cases, the dynamics loss $\ell_d\changeto{}{(x_t,x_{t-1})}$ is typically a
growing function of \changeto{$|x_{t} - F(x_{t-1})|$}{the distance from
$x_t$ to $F(x_{t-1})$}.

The second component of the loss function is an observation loss $\ell_o$.
Given a path $\mbf{p} = (x_1, x_2, \ldots)$, and measurements $\mbf{M} =
(M(\cdot, 1), M(\cdot, 2), \ldots)$, the observation loss function
$\ell_o(\mbf{p}, \mbf{M})$ quantifies how well the path $\mbf{p}$ explains the measurements.
Again, for simplicity, we restrict ourselves to loss functions $\ell_o$ that can be written as:
\[ \ell_o(\mbf{p}, \mbf{M}) = \sum_{t} \ell_o(x_t, M(\cdot, t)) \ . \]
In other words, the observation loss of a path at time $t$ depends only on
its state at time $t$ and the measurements at time $t$. The total loss of a
path $\mbf{p}$ is the sum of its dynamics and observation losses. We note
that the loss of a path depends only on that particular path and the
measurements, and not on the true hidden state. As a result, the loss of
a path can always be evaluated by the algorithm at any given time.

The algorithmic framework we consider in this model is analogous to, and
motivated by the decision-theoretic framework for online
learning~\cite{FS97,CL06}. At time $t$, the algorithm assigns a weight
$w_{\mbf{p}}^t$ to each path $\mbf{p}$ in $\calP$. The estimated state at time $t$ is the weighted mean of the states, where the weight of a state is the total weight of all paths in this state. The loss of the algorithm at time $t$ is the weighted loss of all paths in $\calP$. The theoretical guarantee we look for is that the loss of the algorithm is close to the loss of the best path in $\calP$ in hindsight (or, close to the loss of the top $\epsilon$-quantile path in $\calP$ in hindsight). Thus, if $\calP$ has a small fraction of paths with low loss, and if the loss functions successfully capture the tracking performance, then, the sequence of states estimated by the algorithm will have good tracking performance. 

\section{NormalHedge}

In this section, we describe the NormalHedge algorithm. To present
NormalHedge in full generality, we first need to describe the
decision-theoretic framework for online learning.

The problem of decision-theoretic online learning is as follows. At each
round, a learner has access to a set of $N$ {\em actions}; for our
purposes, an action is any method that
provides a prediction in each round. The learner maintains a distribution
$\pr{i}{t}$ over the action at time $t$. At each time period $t$, each action $i$ suffers a loss  $\ls{i}{t}$ which lies in a bounded range, and the loss of the learner is $\sum_i \pr{i}{t} \ls{i}{t}$. We notice that this framework is very 
general -- no assumption is made about the nature of the actions and 
the distribution of the losses. The goal of the learner is to maintain 
a distribution over the actions, such that its cumulative loss over time 
is low, compared to the cumulative loss of the action with the lowest
cumulative loss.  In some cases, particularly, when the number of experts is very large, we are interested in
acheiving a low cumulative loss compared to the top {\em
$\epsilon$-quantile} of actions. Here, for any $\epsilon$, the top
$\epsilon$-quantile of actions are the $\epsilon$ fraction of actions which
have the lowest cumulative loss.

Starting with the seminal work of Littlestone and Warmuth~\cite{LW94}, the
problem of decision-theoretic online learning has been well-studied
in the literature~\cite{FS97, CFHHSW93,CL06}. The most common algorithm for this
problem is Hedge or Exponential Weights~\cite{FS97}, which assigns to each
action a weight exponentially small in its total loss. In this paper however, we consider a different algorithm
NormalHedge for this problem~\cite{Anon09}, and it is this algorithm that
forms the basis of our tracking algorithm. While the Bayesian averaging
algorithm can be shown to be a variant of Hedge when the loss function is
the log-loss, such is not the case for NormalHedge, and it is a very
different algorithm. A significant advantage of using NormalHedge 
is that it has no 
parameters to tune, yet acheives performance comparable to the best performance
of previous online learning algorithms with optimally tuned parameters.

In the \nhedge\ algorithm, for
each action $i$ and time $t$, we use $\pr{i}{t}$ to denote the {\em \nhedge\ weight}
assigned to action $i$ at time $t$.
At any time $t$, we define the regret $\Rg{i}{t}$ of our algorithm to an
action $i$ as the difference between the cumulative loss of our algorithm
and the cumulative loss of this action. Also, for any real number $x$, we use
the notation $[x]_+$ to denote $\max(0, x)$. The \nhedge\ algorithm is presented below. 
\begin{algorithm}[H]
\caption{NormalHedge}
\label{alg:normalhedge}
\begin{algorithmic}[1]
  \INITIALIZE $\Rg{i}{0} = 0$, $\pr{i}{1} = 1/N$ $\forall i$

  \FOR{$t = 1, 2, \ldots$}
    \STATE Each action $i$ incurs loss $\ls{i}{t}$.
    \STATE Learner incurs loss $\ls{A}{t} = \sum_{i=1}^N \pr{i}{t}
    \ls{i}{t}$.

    \STATE Update cumulative regrets: $\Rg{i}{t} = \Rg{i}{t-1} + (\ls{A}{t}
    - \ls{i}{t})$ $\forall i$.

    \STATE Find $c_t>0$ satisfying
    $\frac{1}{N} \sum_{i=1}^N
    \exp\left(\frac{([\Rg{i}{t}]_+)^2}{2c_t}\right) = e$.

    \STATE Update distribution:
    $\pr{i}{t+1} \propto \frac{[\Rg{i}{t}]_+}{c_t}
    \exp\left(\frac{([\Rg{i}{t}]_+)^2}{2c_t}\right)$ $\forall i$.

  \ENDFOR
\end{algorithmic}
\end{algorithm}
The performance guarantees for the \nhedge\ algorithm, as shown
by~\cite{Anon09} can be stated as follows.
\begin{theorem}
If \nhedge\ has access to $N$ actions, then for all loss
sequences, for all $t$, for all $0 < \epsilon \leq 1$, the regret
of the algorithm to the top $\epsilon$-quantile of the actions is
$O(\sqrt{ t \cdot \ln(1/\epsilon)}+ \ln^2 N)$.
\label{thm:nhedge}
\end{theorem}

Note that the actions which have total loss greater
than the total loss of the algorithm, are assigned zero weight. Since the
algorithm performs almost as well as the best action, in a scenario
where a few actions are significantly better than the rest, the algorithm 
will assign zero weight to most actions.
In other words, the support of the \nhedge\ weights may be a very small
set, which can significantly reduce its computational cost. 

\section{Tracking using NormalHedge}
\label{sec:algorithm}

To apply \nhedge\ directly to tracking, we set each action to be a path in
the state space, and the loss of each action at time $t$ to be the loss of
the corresponding path at time $t$. To make \nhedge\ more robust in a
practical setting, we make a small change to the algorithm: instead of
using cumulative loss, we use a discounted cumulative loss. For a discount
parameter $0 < \alpha < 1$, the discounted cumulative loss of an action $i$
at time $T$ is $\sum_{t=1}^{T} (1 - \alpha)^{T-t} \ls{i}{t}$.
Using discounted losses is common in reinforcement learning~\cite{K96};
intuitively, it makes the tracking algorithm more flexible, and allows it
to more easily recover from past mistakes.

However, a direct application of \nhedge\ is prohibitively expensive in terms of computation cost. Therefore, in the sequel, we show how to derive a Sequential Monte-Carlo based approximation to \nhedge, and we use this approximation in our experiments.

The key observation behind our approximation is that the weights on
actions generated by the \nhedge\ algorithm induce a distribution over the
states at each time $t$.
We therefore use a random sample of states in each round to
approximate this distribution.
Thus, just as particle filters approximate the posterior density on
the states induced by the Bayesian algorithm, our tracking algorithm
approximates the density induced on the states by NormalHedge for tracking.

The main difference between \nhedge\ and our tracking algorithm is that while \nhedge\ always maintains the weights for all the actions, we delete an action from our action list when its weight falls to $0$. We then replace this action by our resampling procedure, which chooses another action which is currently in a region of the state space where the actions have low losses. Thus, we do not spend resources maintaining and updating weights for actions which do not perform well.
Another difference between \nhedge\ and our tracking algorithm is that in our approximation, we do not explicitly impose a dynamics loss on the
actions. Instead, we use a resampling procedure that only considers
actions with low dynamics loss. This also avoids spending resources on actions which have high dynamics loss anyway.

Our tracking algorithm is specified in Algorithm~\ref{alg:tracking}.
Each action $i$ in our algorithm is a path $(x_{i,1}, x_{i,2}, \ldots)$ in
the state space $\X \subset \R^n$.
However, we do not maintain this entire path explicitly for each action;
rather, Step 8 of the algorithm computes $x_{i,t+1}$ from $x_{i,t}$ using
the dynamics function $F$, so we only need to maintain the current state of
each action.
\changeto{(For more details on $F$, see Section~\ref{sec:explanation}.)}
{Recall, applying the dynamics function $F$ should ensure that the path
incurs no or little dynamics loss (see Section~\ref{sec:explanation}).}

We start with a set of actions $\Ev$ initially positioned at states
uniformly distributed over the $\X$, and a uniform weighting over these
actions. 
In each round, like \nhedge, each action incurs a loss determined by its current state,
and the tracker incurs the expected loss determined by the current
weighting over actions.
Using these losses, we update the cumulative (discounted) regrets to each
action. 
However, unlike \nhedge, we then delete all actions with zero or negative regret, and replace them
using a resampling procedure.
This procedure replaces poorly performing actions with actions currently at
high density regions of $\X$, thereby providing a better approximation to
the intended weights.

\begin{figure}
\begin{center}
\begin{tabular}[p]{c}
\begin{minipage}[t]{\textwidth}
\begin{algorithm}[H]
\caption{Tracking algorithm}
\label{alg:tracking}
\begin{algorithmic}[1]
  \INPUT $N$ (number of actions),
  $\alpha$ (discount factor),
  $\Sigma_*$ (resampling parameter)
  $F$ (dynamics function)

  \STATE $\Ev := \{ x_{1,1}, \ldots, x_{N,1}\}$ with $x_{i,1}$ randomly
  drawn from $\X$; $\Rg{i}{0} := 0$; $\pr{i}{0} := 1/N$ $\forall i$

  \FOR{$t = 1, 2, \ldots$}
    \STATE Obtain losses $\ls{i}{t} = \ell_o(x_{i,t})$ for each action $i$
    and update regrets: \\
    $\Rg{i}{t} := (1 - \alpha) \Rg{i}{t-1} + (\ls{A}{t} - \ls{i}{t})$
    where $\ls{A}{t} = \sum_{i=1}^{N} \pr{i}{t-1} \ls{i}{t}$.

    \STATE Delete poor actions: let $X = \{ i : \Rg{i}{t} \leq 0 \}$, set $\Ev := \Ev \setminus X$.

    \STATE Resample actions: $\Ev := \Ev \cup \resample(X, \Sigma_*, t)$.

    \STATE Compute weight of each action $i$:
    $\pr{i}{t} \propto \frac{[\Rg{i}{t}]_+}{c}
    \exp\left(\frac{([\Rg{i}{t}]_+)^2}{2c}\right)$ \\
    where $c$ is the solution to the equation $\frac{1}{N} \sum_{i=1}^{N}
    \exp\left(\frac{([\Rg{i}{t}]_+)^2}{2c}\right) = e$.

    \STATE Estimate: $x_{A,t} := \sum_{i=1}^{N} \pr{i}{t} x_{i,t}$.

    \STATE Update states: $x_{i,t+1} := F(x_{i,t})$ $\forall i$.

  \ENDFOR
\end{algorithmic}
\end{algorithm}
\end{minipage}
\\
\begin{minipage}[t]{\textwidth}
\begin{algorithm}[H]
\caption{Resampling algorithm}
\label{alg:resample}
\begin{algorithmic}[1]
  \INPUT $X$ (actions to be resampled),
  $\Sigma_*$ (resampling parameter),
  $t$ (current time)

  \FOR{$j \in X$}

    \STATE Set $\bar{X} := \{ i : \Rg{i}{t} > 0 \}$.

    \STATE If $\bar{X} = \emptyset$: set $\bar p_i = 1/N$ $\forall i$.
    Else: set $\bar p_i \propto \pr{i}{t-1}$ $\forall i \in \bar{X}$ and $\bar
    p_i = 0$ $\forall i \notin \bar{X}$.

    \STATE Draw $i \sim (\bar p_1,\ldots,\bar p_{N})$.

    \STATE Draw $x_{j,t} \sim \calN(x_{i,t}, \Sigma_*)$, and set $\Rg{j}{t}
    := (1 - \alpha) \Rg{i}{t-1} + (\ls{A}{t} - \ell_o(x_{j,t}))$.

  \ENDFOR
\end{algorithmic}
\end{algorithm}
\end{minipage}
\end{tabular}
\end{center}
\caption{NormalHedge-based tracking algorithm.}
\end{figure}

The resampling procedure is explained in detail in
Algorithm~\ref{alg:resample}.
The main idea is to sample from the regions of the state space with high
weight.
This is done by sampling an action proportional to its weight in the
previous round.
We then choose a state randomly (roughly) from an ellipsoid $\{ x :
(x-x_t)^\top \Sigma_*^{-1} (x-x_t) \leq 1 \}$ around the current state
$x_t$ of the selected action; the new action inherits the history of the
selected action, but has a current state which is different from (but close
to) the selected action.
This latter step makes the new state distribution smoother than the one in
the previous round, which may be supported on just a few states if only a
few actions have low losses.
We note that $\Sigma_*$ can be set so that the resampling procedure only
samples actions with low dynamics loss (and Step 8 of the algorithm ensures
that the remaining actions in the set $\Ev$ do not incur any dynamics
loss)\changeto{.}{; thus, our algorithm does not explicitly compute a
dynamics loss for each action.}

\section{Simulations}

For our simulations, we consider the task of tracking an object in a
simple, one-dimensional state space. To evaluate our algorithm, we measure the
distance between the estimated states, and the true states of the object.

Our experimental setup is inspired by the application of tracking faces in
videos, using a standard face detector~\cite{VJ01}.
In this case, the state is the location of a face, and each measurement
corresponds to a score output by the face detector for a region in the
current video frame.
The goal is to predict the location of the face across several video
frames, using these scores produced by the detector.
The detector typically returns high scores for several regions around the
true location of a face, but it may also erroneously produce high scores
elsewhere.
And though in some cases the detection score may have a probabilistic
interpretation, it is often difficult to accurately characterize the
distribution of the noise.

\begin{figure}
\begin{center}
\begin{tabular}{cc}
\includegraphics[width=0.45\textwidth]{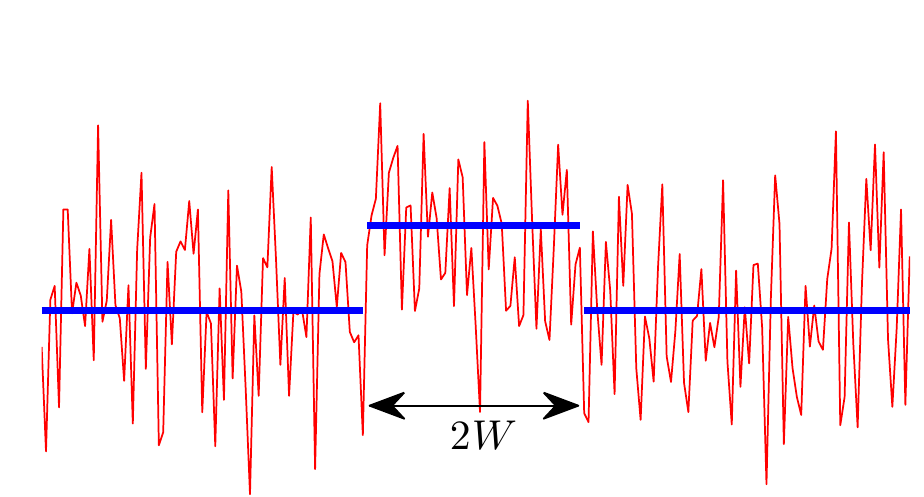} &
\includegraphics[width=0.45\textwidth]{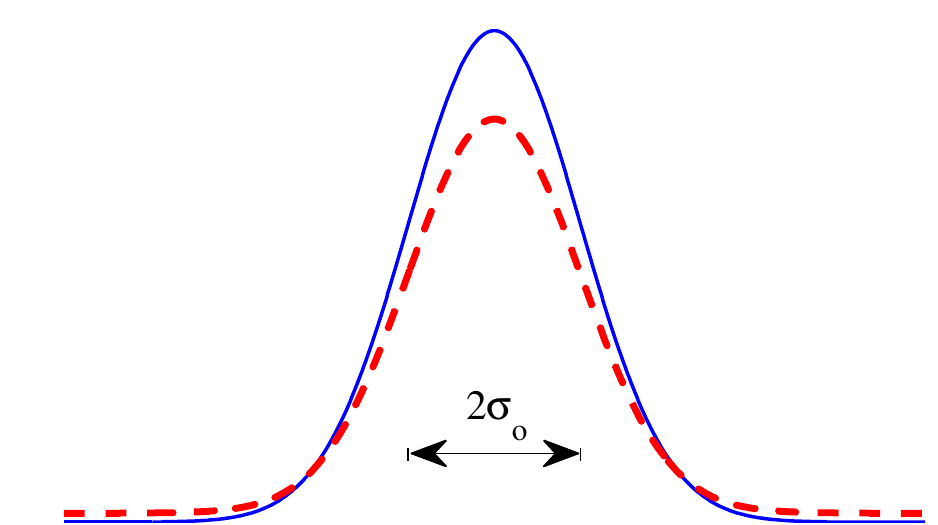} \\
Blue: $H(x,z_t)$, Red: $M(x,t)$. &
Blue: $\rho = 0$, Red: $\rho = 0.2$.
\end{tabular}
\end{center}
\caption{
Plots of the measurements (as a function of $x$) for $\rho=0$ and $\sigma_o
= 1$ and the density of the
noise $n_t(x)$.
}
\label{fig:expt-diagrams}
\end{figure}

The precise setup of our simulations is as follows.
The object to be tracked remains stationary or moves with velocity at most
$1$ in the interval $[-500,500]$.
At time $t$, the true state is the position $z_t$; the measurements
correspond to a $1001$-dimensional vector $\mbf{M}(t) = [ M(-500, t),
M(-499,t), \ldots, M(499,t),M(500,t)]$ for locations in a grid $G = \{-500,
-499, \ldots, 499, 500\}$, generated by an additive noise process
\[ M(x,t) = H(x, z_t) + n_t(x). \]
Here, $H(x, z_t)$ is the square pulse function of width $2W$ around the true state
$z_t$: $H(x,x_t) = 1$ if $|x - z_t| \leq W$ and $0$ otherwise (see
Figure~\ref{fig:expt-diagrams}, left).
The additive noise $n_t(x)$ is
randomly generated independently for each $t$ and each $x \in G$, using the
mixture distribution
$$ (1-\rho) \cdot \calN(0, \sigma_o^2) + \rho \cdot
\calN(0, (10\sigma_o)^2)$$
(see Figure~\ref{fig:expt-diagrams}, right).
The parameter $\sigma_o$ represents how noisy the measurements are
relative to the signal, and the parameter $\rho$ represents the fraction of
outliers.
In our experiments, we fix $W = 50$ and vary $\sigma_o$ and $\rho$.
The total number of time steps we track for is $T = 200$.

In the generative framework, the dynamics of the object is represented by the
transition model $x_{t+1} \sim \calN(x_{t}, \sigma_d^2)$, and the
observations are represented by the measurement process $M(x, t) \sim
\calN( H(x, x_t), \sigma_o^2)$. Thus, when $\rho = 0$, the observations are
generated according to the measurement process supplied to the generative
framework; for $\rho > 0$, a $\rho$ fraction of the observations are
outliers.

For the \explanatory\ framework, the expected state dynamics function $F$ is the identity function,
and the observation loss of a path $\mbf{p} = (x_1, x_2, \ldots )$ at time
$t$ is given by
\[ \ell_o(x_t,M(\cdot,t)) = -\sum_{x \in [x_t-W, x_t+W] \cap G} q(M(x, t)) \]
where $q(y) = \min(1 + \sigma_o, \max(y, -\sigma_o))$ clips the
measurements to the range $[-\sigma_o, 1 + \sigma_o]$.
That is, the observation loss for $x_t$ with respect to $M(\cdot,t)$ is the
negative sum of thresholded measurement values $q(M(x,t))$ for $x$ in an
interval of width $2W$ around $x_t$.

Given only the observation vectors $\mbf{M}$, we use three different
methods to estimate the true underlying state sequence $(z_1, z_2, \ldots)$.
The first is the Bayesian algorithm, \changeto{computed over the
discretization $G$.}{which recursively applies Bayes' rule to update a
posterior distribution using the transition and observation model.
The posterior distribution is maintained at each location in the
discretization $G$.}
For the Bayesian algorithm, we set $\sigma_o$ to the actual value of
$\sigma_o$ used to generate the observations,
and we set $\sigma_d = 2$.
The value of $\sigma_d$ was obtained by tuning on measurement vectors
generated with the same true state sequence, but with independently generated
noise values.
The prior distribution over states assigns probability one to the true
value of $z_1$ (which is $0$ in our setup) and zero elsewhere.
The second algorithm is our algorithm (NH) described in
Section~\ref{sec:algorithm}. For our algorithm, we use the parameters
$\Sigma_* = 400$ and $\alpha = 0.02$. These parameters were also obtained
by tuning over a range of values for $\Sigma_*$ and $\alpha$.
We also compare our algorithm with the particle filter (PF), 
which uses the same parameters as with the Bayesian algorithm, and predicts
using the expected state under the (approximate) posterior distribution.
For our algorithm, we use $N=100$ actions, and for the particle filter, we use $N=100$ particles.
For our experiments, we use an implementation of the particle filter due to~\cite{dF00}.

\begin{figure*}
\begin{center}
\begin{tabular}{ccc}
\includegraphics[height=0.123\textheight]{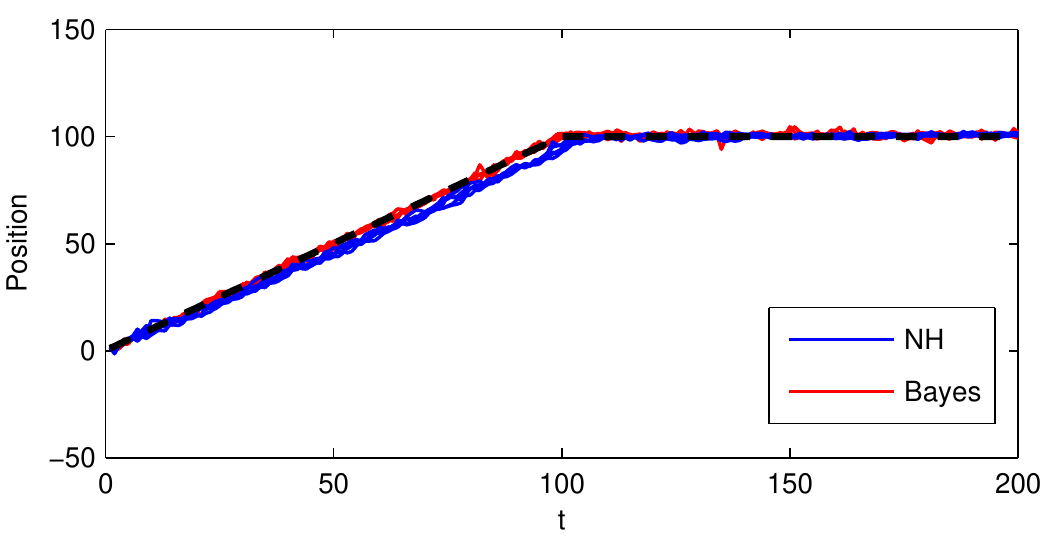} &
\includegraphics[height=0.123\textheight]{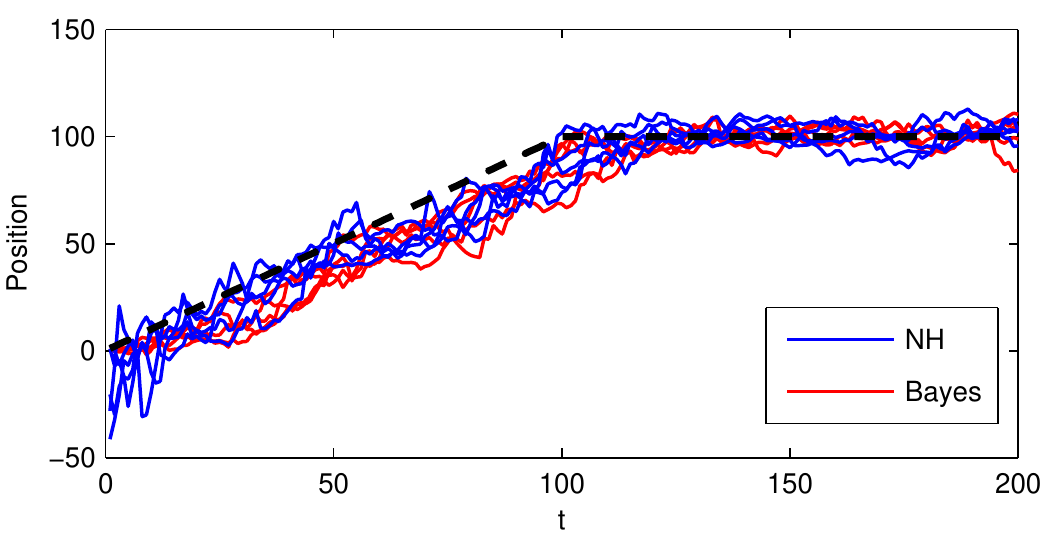} \\
$\sigma_o = 1$, $\rho = 0$ & $\sigma_o = 8$, $\rho = 0$ \\
\includegraphics[height=0.123\textheight]{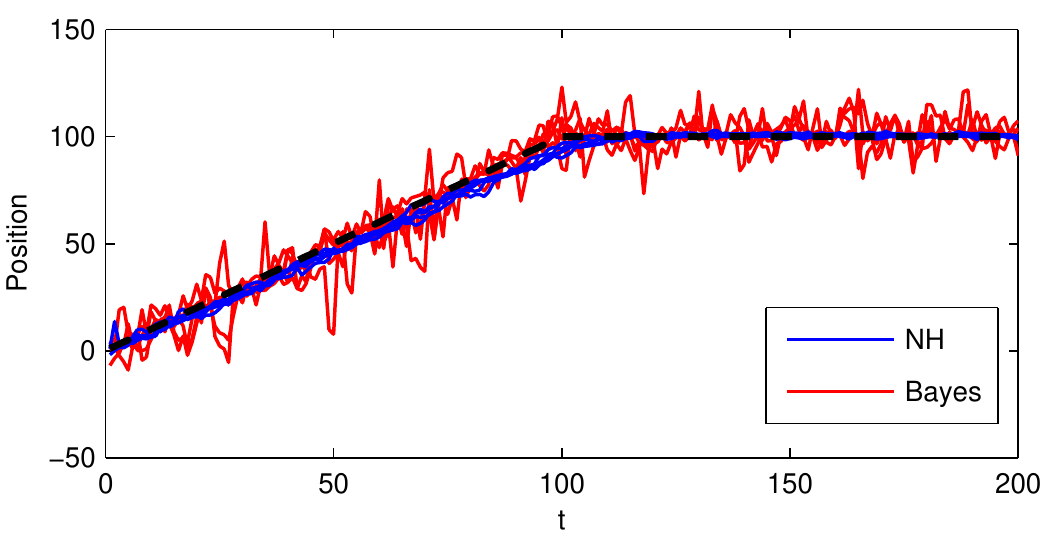} &
\includegraphics[height=0.123\textheight]{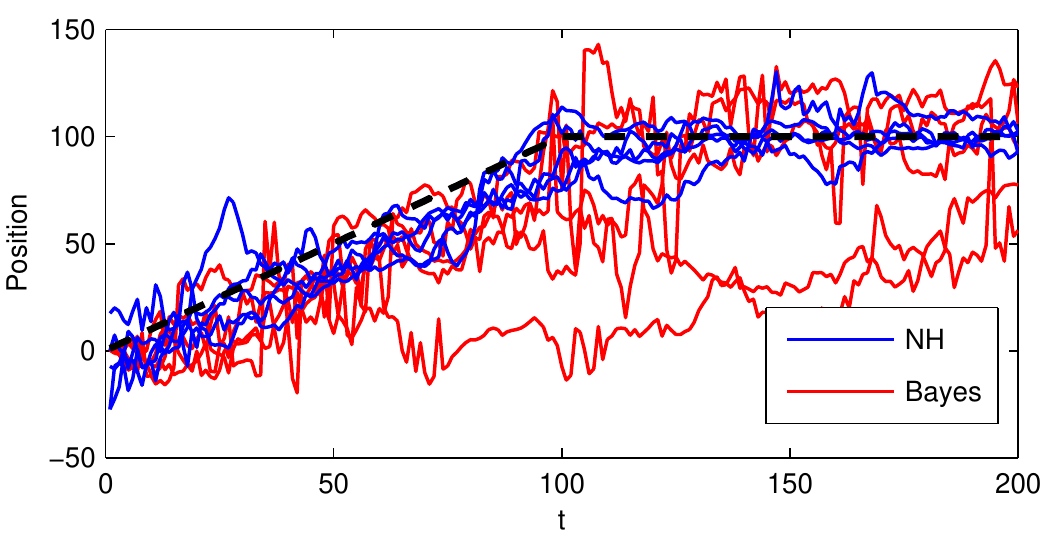} \\
$\sigma_o = 1$, $\rho = 0.1$ & $\sigma_o = 8$, $\rho = 0.1$ \\
\includegraphics[height=0.123\textheight]{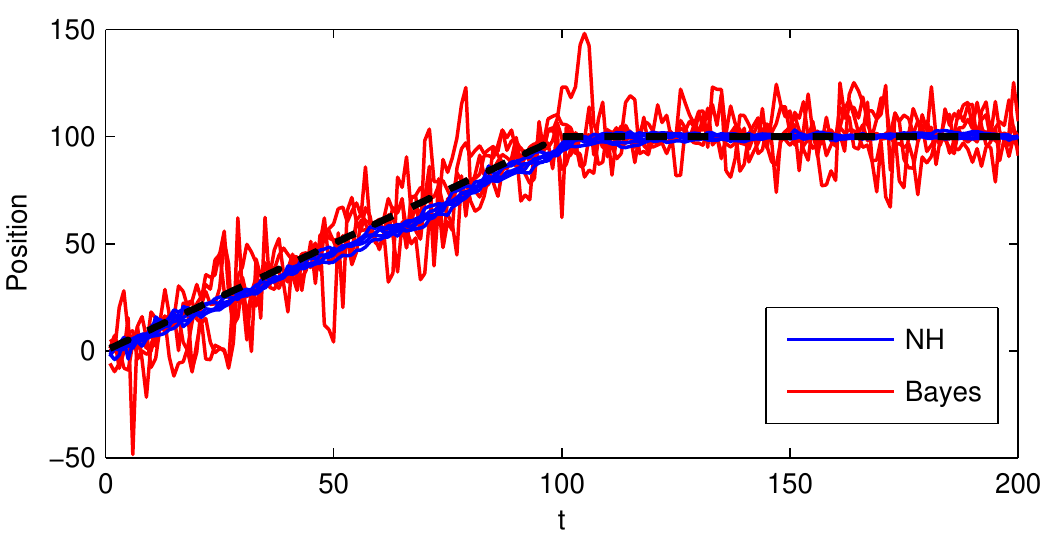} &
\includegraphics[height=0.123\textheight]{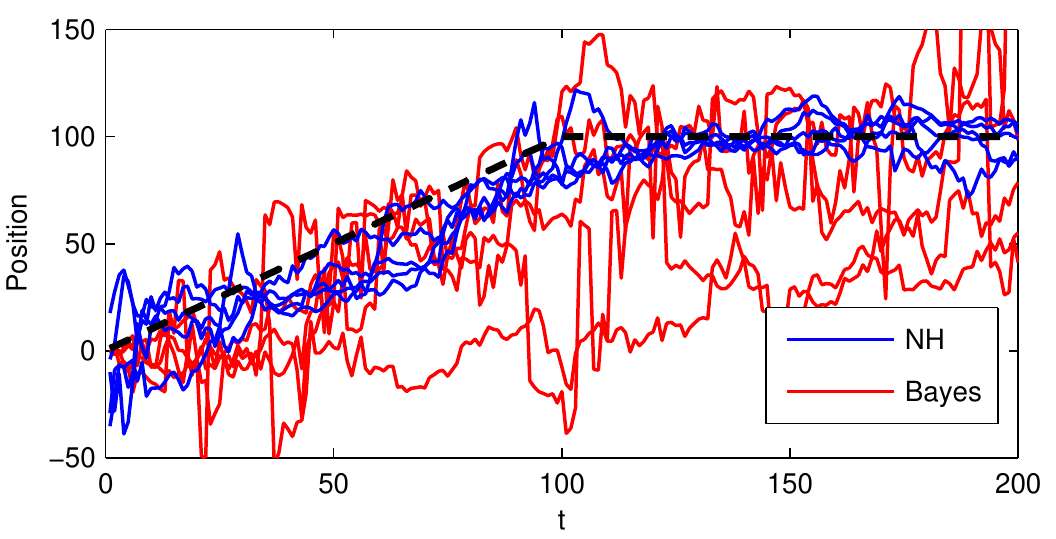} \\
$\sigma_o = 1$, $\rho = 0.2$ & $\sigma_o = 8$, $\rho = 0.2$
\end{tabular}
\end{center}
\caption{
Predicted paths in five simulations.
First column: low noise ($\sigma_o = 1$).
Second column: high noise ($\sigma_o = 8$).
The blue lines correspond to our algorithm,
the red lines correspond to the Bayesian algorithm,
and the dashed black line represents the true states.}
\label{fig:sims}
\end{figure*}

Figure~\ref{fig:sims} shows the true state and the states predicted by our
algorithm (Blue) and the Bayesian algorithm (Red) for two different values
of $\sigma_o$ for $5$ independent simulations.
Table~\ref{tab:tabres} summarizes the performance of these algorithms for different values of the parameter $\rho$, for two different values of the noise parameter $\sigma_o$.
We report the average and standard deviation of the RMSE
(root-mean-squared-error) between the true state and the predicted state.
The RMSE is computed over the $T=200$ state predictions for a single
simulation, and these RMSE values are averaged over $100$ independent
simulations.

Our experiments show that the Bayesian algorithm performs well when
$\rho=0$, that is, it is supplied with the correct noise model; however,
its performance degrades rapidly as $\rho$ increases, and becomes very poor
even at $\rho=0.2$. On the other hand, the performance of our algorithm
does not suffer appreciably when $\rho$ increases. The degradation of
performance of the Bayesian algorithm is even more pronounced, when the
noise is high with respect to the signal ($\sigma_o = 8$). The particle filter
suffers a even higher degradation in performance, and has poor performance
even when $\rho=0.01$ (that is, when \changeto{$99$-percent}{$99\%$} of the observations are
generated from the correct likelihood distribution supplied to the particle
filter). Our results indicate that the Bayesian algorithm is very
sensitive to model mismatches. On the other hand, our algorithm, when
equipped with a clipped-loss function, is extremely robust to model mismatches. In particular, our algorithm provides a RMSE value of $19.6$ even under high noise ($\sigma_o = 8$), when $\rho$ is as high as $0.4$.

Some additional experiments with our algorithm are included in the
supplementary appendix; they illustrate how the performance of
our algorithm varies with the parameters $\Sigma_*$ and $\alpha$, and
tabulates the performance of our algorithm for higher values of $\rho$.

\begin{table}
\begin{center}
\caption{Experimental Results.
Root-mean-squared-errors of the predicted positions over $T=200$
time steps for our algorithm (NH), the Bayesian algorithm, and the
particle filter (PF).
The reported values are the averages and standard deviations over $100$
simulations.
\label{tab:tabres}
}
\vspace{0.3cm}
\begin{small}
{
\renewcommand{\arraystretch}{1.5}
\renewcommand{\tabcolsep}{0.12cm}
\begin{tabular}[p]{cc}
Low Noise ($\sigma_o = 1$) & High Noise ($\sigma_o = 8$) \\
\begin{tabular}{|c|r@{.}l@{ $\pm$ }l|r@{.}l@{ $\pm$ }l|r@{.}l@{ $\pm$ }l|}
\cline{1-10}
$\rho$
& \multicolumn{3}{c|}{NH}
& \multicolumn{3}{c|}{Bayes}
& \multicolumn{3}{c|}{PF} \\
\cline{1-10}
\hline
$0.00$
&  3&18 & $0.33$
&  1&17 & $0.09$
&  1&23 & $0.11$
\\
$0.01$
&  3&21 & $0.34$
&  1&90 & $0.25$
&  3&98 & $1.06$
\\
$0.05$
&  3&26 & $0.34$
&  3&99 & $0.52$
& 81&70 & $1.74$
\\
$0.10$
&  3&31 & $0.35$
&  6&40 & $0.84$
& 81&70 & $1.74$
\\
$0.15$
&  3&42 & $0.34$
&  8&38 & $1.10$
& 81&70 & $1.74$
\\
$0.20$
&  3&52 & $0.41$
& 10&28 & $1.24$
& 81&70 & $1.74$
\\
\hline
\end{tabular}
&
\begin{tabular}{|c|r@{.}l@{ $\pm$ }l|r@{.}l@{ $\pm$ }l|r@{.}l@{ $\pm$ }l|}
\cline{1-10}
$\rho$
& \multicolumn{3}{c|}{NH}
& \multicolumn{3}{c|}{Bayes}
& \multicolumn{3}{c|}{PF} \\
\cline{1-10}
\hline
$0.00$
&  10&93 & $2.52$
&  10&98 & $2.33$
&  14&35 & $5.16$
\\
$0.01$
&  11&26 & $3.43$
&  12&76 & $3.07$
&  44&29 & $16.7$
\\
$0.05$
& 12&03 & $3.47$
& 19&75 & $6.70$
& 81&70 & $1.74$
\\
$0.10$
&  12&25 & $2.93$
&  27&33 & $10.9$
& 81&70 & $1.74$
\\
$0.15$
&  13&38 & $3.07$
&  32&78 & $13.1$
& 81&70 & $1.74$
\\
$0.20$
&  14&15 & $3.88$
&  43&99 & $26.8$
& 81&70 & $1.74$
\\
\hline
\end{tabular}
\end{tabular}
}
\end{small}
\end{center}
\end{table}

\section{Related work}

The generative approach to tracking has roots in control and estimation
theory, starting with the seminal work of Kalman~\cite{Kal60}. The most
popular generative method used in tracking is the particle
filter~\cite{DdFG01}, and its numerous variants.
The literature here is vast, and there have been many exciting
developments in recent years (\emph{e.g.}~\cite{vdMDdFW00,KdFD05}); we
refer the reader to \cite{DJ08} for a detailed survey of the results.

The suboptimality of the Bayesian algorithm under model mismatch has been
investigated in other contexts such as classification~\cite{Dom00,GL07}.
The view of the Bayesian algorithm as an online learning algorithm for
log-loss is well-known in various communities, including information theory
/ MDL~\cite{MF98,Grunwald07} and computational learning
theory~\cite{F00,KN00}. \changeto{(\emph{e.g.}~\cite{HW98,KdR08})}{} In our work, we look beyond the Bayesian algorithm and log-loss to consider other
loss functions and algorithms that are more appropriate for our task.

There has also been some work on tracking in the online learning literature
(see, for example,~\cite{HW98, KdR08}); there, however, they study a very
different model for tracking.

\section{Conclusions}

In this paper, we introduce an \explanatory\ framework for tracking based
on online learning, which broadens the space for designing algorithms that
need not conform to the standard Bayesian approach to tracking.
We propose a new algorithm for tracking in this framework that deviates
significantly from the Bayesian approach.
Experimental results show that our algorithm significantly outperforms the
Bayesian algorithm, even when the observations are generated by a
distribution deviating just slightly from the model supplied to the
Bayesian algorithm.
Our work reveals an interesting connection between decision theoretic
online learning and Bayesian filtering.

\subsubsection*{References}
{\def\section*#1{}\small \bibliography{paper} \bibliographystyle{unsrt}}

\end{document}